\documentclass[11pt]{article} 
\usepackage{rldmsubmit,palatino}
\usepackage{cite}
\usepackage{hyperref}
\usepackage{amsmath,amssymb,amsfonts}
\usepackage{algorithmic}
\usepackage{graphicx}
\usepackage[caption=false]{subfig}
\usepackage{textcomp}
\usepackage{gensymb}
\usepackage{siunitx}
\usepackage{todonotes}

\todostyle{}{color=red!40}

\title{Multi-Target Radar Search and Track Using Sequence-Capable Deep Reinforcement Learning}

\author{Jan-Hendrik Ewers \\
PxCDT \\
Leonardo Electronics\\
Edinburgh, Scotland \\
\href{https://orcid.org/0000-0002-8498-1410}{0000-0002-8498-1410}
\And
David Cormack \\
PxCDT \\
Leonardo Electronics\\
Edinburgh, Scotland \\
david.cormack03@leonardo.com
\And
Joe Gibbs \\
Autonomous Systems and Connectivity \\
University of Glasgow\\
Glasgow, Scotland \\
j.gibbs.1@research.gla.ac.uk
\And
David Anderson \\
Autonomous Systems and Connectivity \\
University of Glasgow\\
Glasgow, Scotland \\
dave.anderson@glasgow.ac.uk
}

\begin{document}

\maketitle

\begin{abstract}
The research addresses sensor task management for radar systems, focusing on efficiently searching and tracking multiple targets in a complex 3D environment. The study proposes a reinforcement learning solution that balances surveillance and tracking to maximize information gathering. The approach develops a 3D simulation environment using an active electronically scanned array radar with a fixed field of view. A multi-target tracking algorithm using an Unscented Kalman filter improves observation data quality by maintaining a reliable track list of target states between scans. Three neural network architectures were compared: a simple flattening approach, a bidirectional gated recurrent units method, and a combined approach using recurrent units with multi-headed self-attention. Two pre-training techniques were applied: behavior cloning to approximate a random search strategy and an auto-encoder to pre-train the track list feature extractor. Experimental results revealed that search performance was relatively consistent across most methods. The real challenge emerged in simultaneously searching and tracking targets. The multi-headed self-attention architecture demonstrated the most promising results, highlighting the potential of sequence-capable architectures in handling dynamic tracking scenarios. The research provides a robust framework for improving radar system performance by optimizing the balance between searching and tracking. By demonstrating how reinforcement learning can effectively manage sensor tasks, the study offers insights into more intelligent and adaptive radar systems. The key contribution lies in showing how machine learning techniques can overcome traditional limitations in sensor management, potentially improving how radar systems identify and track multiple targets in complex environments.
\end{abstract}


\acknowledgements{
    Jan-Hendrik Ewers is a PhD candidate on secondment at Leonardo Electronics funded and is supported by the Engineering and Physical Sciences Research Council  grant number EP/T517896/1-312561-05. Joe Gibbs is a sponsored PhD candidate with Leonardo Electronics.
}

\startmain
\section{Introduction}
Active Electronically Scanned Array (AESA) RADAR systems deployed to fighter aircraft are required to perform surveillance (search) and tracking (revisit) \cite{white2008radar} to scan for new targets and update existing track information for use by the pilot. Determining the task that will generate the most information is referred to as sensor management \cite{hero2011sensor}.

Reinforcement learning (RL) is a machine learning approach used to train an autonomous agent to generate a set of actions $a_{k}$ based on information contained within an observation state $s_{k}$ and a reward calculated based on the impact of the policy on the environment. RL has been employed in many domains such as drone racing \cite{Kaufmann2023} or wilderness search and rescue \cite{Ewers2024} with RL outperforming state of the art algorithms. A common reason is that RL is able to learn underlying patterns within the data that traditional methods cannot. Within radar task management, RL has been applied to the sensor management problem before. 
\cite{Shaghaghi2018} proposes a RL-based approach with monte carlo tree search whilst \cite{Qu2019} uses earliest start time with RL. However, both only consider the tracking portion and neither implement a full Multi-Target Tracking (MTT) simulation. \cite{Xiong2023} applies multi agent RL to a 2D simulation but also does not consider the search aspect. Therefore this paper proposes a RL solution to the problem in a full 3D MTT environment for both search and tracking.



\section{Methodology}
\label{sect:method}
\subsection{Simulation and MTT Feature Extraction}
The agent is trained in a basic but representative environment simulating a typical engagement with a observer platform scanning an area of interest. The RADAR is assumed to generate measurements in a field-of-view of $9\degree$. Since the sensor is assumed to be an AESA RADAR, the agent actions are considered to occur instantaneously. A fixed number of targets are simulated with all targets birthed at $t=t_{0}$ with deaths at $t=t_{f}$. A simulation time step of $50\text{ms}$ with the sensor generating observations at the same frequency as the MTT algorithm for simplicity. The sensor management agent utilises knowledge of current tracks in the area-of-interest as observations to determine the agent policy. In the simplest case, the agent could take observations from RADAR returns. However, there are several limitations of this approach. RADAR observations are noise-corrupted and will contain clutter or anomalies which will degrade agent performance and RADAR returns will provide information only when a sector is scanned, with no method of maintaining target information between sector revisits. To combat this, a MTT algorithm is implemented as part of the feature extraction process improve the quality of the observation data, by populating a track list of kinematic target states at the current timestep. The MTT algorithm is implemented using the open-source Stone Soup package \cite{barr2022stone} using an Unscented Kalman filter estimator \cite{julier2004unscented} for performing measurement updates. Track \textit{assignment} is performed with the nearest neighour algorithm using the 2D Munkres assignment algorithm and the Mahalanobis distance metric as the hypothesiser to populate the cost matrix. It is assumed that the targets birth at the start of the simulation and live throughout the duration of the scenario. Tracks are deleted from the track list if the Frobenius norm of the target covariance is above a set threshold. A constant-velocity target model is used as the predictor component with a state vector $\hat{\mathbf{x}}_{k}=\begin{bmatrix}x&\dot{x}&y&\dot{y}&z&\dot{z}\end{bmatrix}^{\top}\in\mathbb{R}^{6}$.

\subsection{Scan History}

This work employs a probabilistic approach to model the evolution of scan values over time. Each scan is represented by a bivariate Gaussian distribution with zero correlation;
\begin{equation}
    p(\vec x, \vec \mu, \vec \sigma) = \frac{1}{2\pi\sigma_a\sigma_b} \exp{\left[ -\frac{1}{2}\left(
        \left(\frac{x_a-\mu_{a}}{\sigma_a}\right)^2 +  \left(\frac{x_b-\mu_{b}}{\sigma_b}\right)^2 
    \right)
    \right]} 
    \label{eq:bivariate_normal_distribution}    
\end{equation}
The initial scan uncertainty is characterized by a standard deviation $\sigma_0$ determined to encompass 95\% of the probability mass within the field of view (FoV). 

To model the gradual loss of confidence in older scans, the standard deviation is diffused over time by $\sigma_{t} = \sigma_{0} (1+\gamma)^t$
where $\gamma$ is a parameter controlling the diffusion rate. The maximum possible scan value at any given time step is $p_{\textrm{max},T} = \sum^T_{t=0} (2\pi\sigma_0^2(1+\gamma)^{2t})^{-1}$.

Scan history is maintained using a first-in, first-out (FIFO) queue of length $N_{SV}$. Each entry in the queue represents a single scan and consists of its mean value ($\mu = [\psi, \theta]$) and the time step at which it was acquired. The scan value at any point in the search space is computed by summing the contributions of all scans in the queue, weighted by their respective uncertainties and time-dependent diffusion. 

Finally, the scan value is normalized by the maximum possible scan value to obtain a scaled scan value (SSV), facilitating comparison across different time steps and search areas.

\subsection{Action Space}

The agent's action space is discrete, consisting of a grid of scan directions. The number of actions per axis is determined by the ratio of the instantaneous field of view (IFOV) to the total field of view (FOV), accounting for the circular sensor beam. This guarantees full coverage of the search space while introducing controlled overlap between actions.


\subsection{Observation Space}
\label{sect:method_observation_space}

The observation space comprises two key components: the track list ($s_\mathrm{TL}$) and the scan history ($s_\mathrm{SH}$). The track list is a dynamic representation of all currently identified tracks. Each track within the list is characterized by its position, velocity, and a measure of confidence. The position is expressed in polar coordinates (range, azimuth, elevation), while velocity is given in Cartesian coordinates. The confidence metric reflects the robustness of the track estimate. The scan history ($s_\mathrm{SH}$) is a rasterized representation of the sensor's scan data, visualized as a two-dimensional image. This image captures the sensor's measurements over a specific field of view, providing valuable contextual information about the environment.

\subsection{Rewards}

Minimizing the scan value implies better search performance giving $r_{SV,t} = - SSV([\psi, \theta], t)$.
The tracking performance is given by the reduction in covariance across all the tracks with detections. This further rewards efficient tracking by detecting multiple targets. This is calculated using
$
    r_{TL,t} = \sum \left\{ 
    ||P_t||_\mathrm{fro} - ||P_{t-1}||_\mathrm{fro} 
    ~|~
   D_t \bigcup D_{t-1} 
   \right\}
$
where $D_t$ is the set of detections by the sensor at timestep $t$ and $P_t$ is the covariance matrix of the associated track for $D_t$.
The total reward is then the summation of reward components; $r_t =  r_{SV,t} + r_{TL,t}$.

\subsection{Network Design}
\label{sect:network_design}

$s_{TL}$ is a varying sized matrix with the number of rows depending on the number of currently detected tracks. For non-temporal data structures, the use of bidirectional recurrent architecture has shown promising results \cite{Wang2020} with the Gated Regression Unit (GRU) being one of the best variants for lower complexity data. Recent developments, however, have been moving away from recurrent architectures in favour of transformers \cite{Vaswani2017} where the revolution lies in the Multi-Headed Self-Attention (MHSA) block that significantly accelerates training times. We therefore propose three architectures: PPO-Flat, PPO-BiGRU, PPO-BiGRU+MHSA.
BiGRU+MHSA still requires the use of a recurrent block to handle the sequence-to-vector encoding because MHSA only performs sequence-to-sequence operations. The core policy network is a Multi-Layer Perceptron (MLP) which requires a fixed-size input. However, this part of the architecture is only $50\%$ of the size of the standalone BiGRU architecture. Flat is a simple row-major matrix transformation from 2D to 1D with zero-value padding to account for the varying rt.
To effectively handle $s_{SH}$, we use the NatureCNN architecture from \cite{Mnih2015} which has shown excellent performance in image-based operations for reinforcement learning. 
The final latent-space encodings $z_{TL}$ and $z_{SH}$ are then concatenated into $z$ before being fed into the MLP core-network. 


\subsubsection{Behaviour Cloning}
\label{sect:behaviour_cloning}

Imitation learning (IL) is a powerful technique when coupled with RL. IL is commonly referred to as offline RL due to its similarities. In IL a dataset is used to train a policy whereas in RL this dataset is created online through interaction with the environment. The most ubiquitous IL method is behaviour cloning (BC) \cite{Foster2024} which collapses IL down to a supervised learning problem. In BC the states $s \in S$ are inputs to the policy with actions $a \in A$ being the labels. During training, the mean-squared loss between the estimate action $\hat a$ and the true action $a$ is minimized until $\pi_\mathrm{student} \approx \pi_\mathrm{teacher}$.

Radar search is a simple task which can be solved using a random policy. This gives good search performance but has no ability for tracking. However, this can still be used to pre-train the RL policy to ensure it is good at searching to begin with and can thus quickly detect targets potentially reducing training time. Therefore the RL policy tries to approximate the random policy such that $\pi_\theta \approx \pi_\mathrm{random}$ where $\pi_\mathrm{random} \sim \mathcal{U}(0,N_a)^{1 \times 2}$.
Without BC, the policy would first have to learn how to search before being able to receive the sparse rewards for correctly tracking. 

\subsubsection{Auto-encoder}
\label{sect:auto-encoder}
\autoref{sect:behaviour_cloning} outlines how the network is pre-trained using BC but only to maximize the search performance. This means that the track list feature extractor is left untrained. Thus, to also maximize the potential for tracking performance during RL, the aforementioned feature extractor is pre-trained using the Auto-Encoder (AE) paradigm \cite{Sarankumar2024}. The AE is a popular unsupervised training method which can train an encoder, or feature extractor in this case, to maximize the information throughput during encoding. The AE is has two components: the encoder $f_E$ and the decoder $f_D$. These are used to approximate $x$ with $\hat x = f_D(f_E(x))$. The mean squared error $\frac{1}{n} \sum^n_{i=0} (\hat x_i - x_i)^2$ is then used to minimze the difference between the true and approximate values. In this work, $x \sim \mathcal{U}(-1,1)^{N_\mathrm{track,max}\times7}$ which is inline with the observation space definition of the track list in \autoref{sect:method_observation_space} resulting in a track list feature extractor that is able to encode the observation to a meaningful latent space.

\section{Results}
\label{sect:results}

\begin{figure*}[tb!]
    \centering    
    \subfloat[GOSPA distance is the overall metric with lower values being better.\label{fig:results_gospa_distance}]{\includegraphics[width=0.24\linewidth]{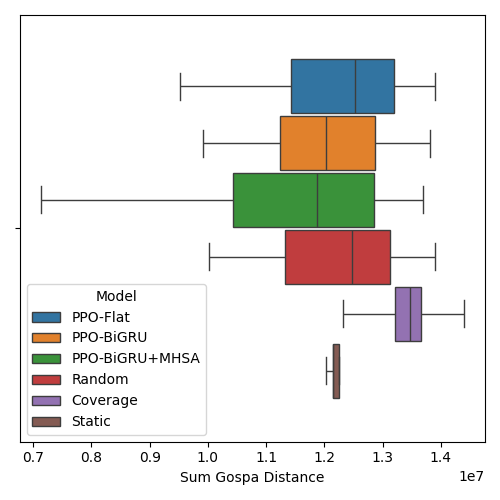}}
    \hfill
    \subfloat[GOSPA localisation is the tracking error score if a truth and track can be associated.\label{fig:results_gospa_localisation}]{\includegraphics[width=0.24\linewidth]{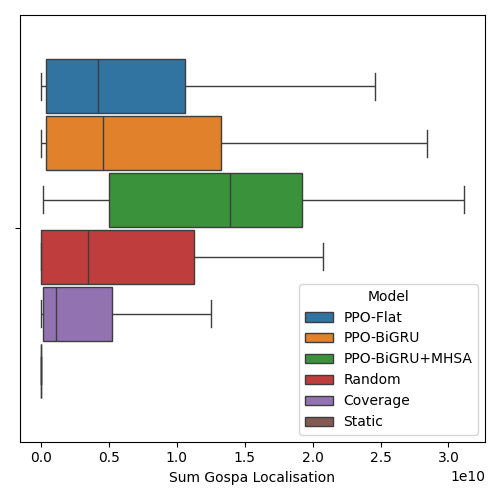}}
    \hfill
    \subfloat[GOSPA false measures the amount of tracks without an associated ground truth.\label{fig:results_gospa_false}]{\includegraphics[width=0.24\linewidth]{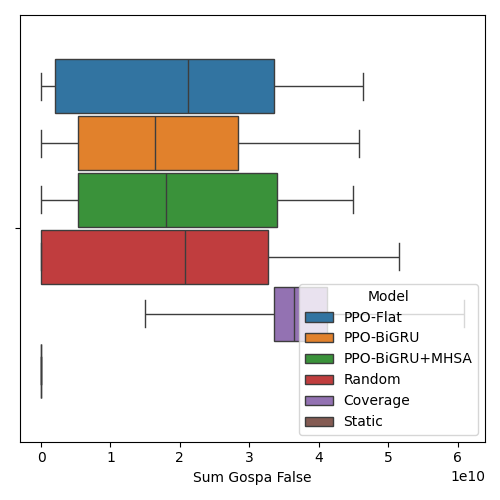}}
    \hfill
    \subfloat[GOSPA missed measures the amount of ground truths without an associated track.\label{fig:results_gospa_missed}]{\includegraphics[width=0.24\linewidth]{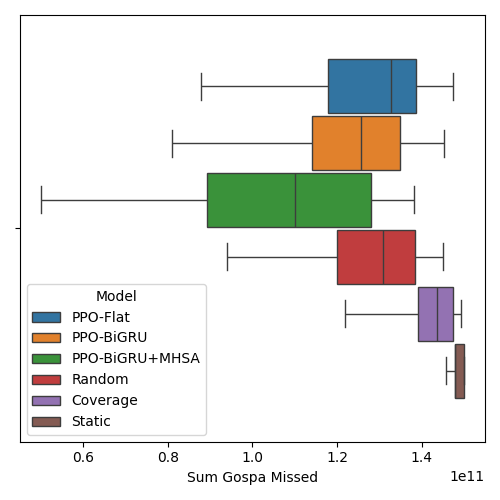}}
    \caption{GOSPA distance (\autoref{fig:results_gospa_distance}) and three of its components \cite{Rahmathullah2017}. GOSPA switching has been excluded as algorithms had a value of $0$ for this metric showing that any detected track was never falsely associated to a different ground truth throughout its existence.}
    \label{fig:results_gosap_all}
\end{figure*}


Three generic baseline policies are implemented for comparison: Static, where an initial action is sampled uniformly from the action space and then held constant throughout the episode; Random, where an action is sampled uniformly from the action space at each timestep; and Coverage, which scans each line sequentially. The models were trained on a \texttt{Intel i7-9700} with $32\si{\giga\byte}$ of RAM and no dedicated GPU. Each model was trained twice over $\SI{1e7}{}$ steps with early termination if no improvement was recorded in an evaluation environment after $\SI{25e5}{}$ steps. After training each model was analysed over $100$ full simulations to collect the summary data. The core policy is of size $2 \times 128$ and NatureCNN outputs a feature of dimension $1\times128$ across all variations. BiGRU has a hidden dimension of $64$, whilst this is $32$ for the BiGRU aspect of BiGRU+MHSA. Both BiGRU variants output a feature dimension of $1\times64$ with flat having a feature dimension of $1\times 120$ due to the lack of any reduction in observation.



In simulation, the 'Static' approach exhibited the lowest mean search reward at $\SI{-2.52E+01}{} \pm \SI{6.48E+00}{}$, indicating significant under-performance. Conversely, the 'Coverage' method demonstrated the highest average search reward at $\SI{-5.02E-02}{} \pm \SI{3.44E-02}{}$. Notably, the three PPO-based architectures and 'Random' exhibited closely clustered mean search rewards from $\SI{-1.37E-01}{}$ to $\SI{-1.27E-01}{}$, suggesting comparable search performance among these five algorithms.

These results highlight that the search component of the search-and-track problem itself is not inherently challenging. The substantial underperformance of 'Static' can be attributed to its lack of search behavior, rendering it the least optimal solution for this task. It is evident from results shown previously that search is not the difficult aspect of this problem. Being able to search and track simultaneously, however, is what is important. Static, without any tracking capabilities, has the highest mean covariance value $\SI{2.63E+03}{} \pm \SI{2.36E-02}{}$. Random on the other hand has a lower value of $\SI{9.48E+02}{} \pm \SI{9.22E+02}{}$ but the largest standard deviation. This is expected since the policy has a chance of perfectly tracking a target as well as never tracking it due to the randomness. Coverage on the other hand performs similarly to PPO-Flat and PPO-BiGRU with values of $\SI{5.74E+02}{} \pm \SI{4.25E+02}{}$, $\SI{7.27E+02}{} \pm \SI{8.26E+02}{}$, and $\SI{7.65E+02}{} \pm \SI{7.76E+02}{}$ respectively. However, the $5^{th}$ percentile whisker for coverage is higher than all PPO algorithms showing that there is potential for more performance there. The best algorithm, is however PPO-BiGRU+MHSA with a mean covariance norm value  of $\SI{3.91E+02}{} \pm \SI{4.43E+02}{}$. 

The General Optimal Sub-Pattern Association (GOSPA) \cite{Rahmathullah2017} is a summary metric of various tracking criteria. The overall score, GOSPA distance, can be seen in \autoref{fig:results_gospa_distance} with static unexpectedly performing well. However, this is explained by \autoref{fig:results_gospa_localisation} and \autoref{fig:results_gospa_false} which show that static does not track. Therefore the overall score is low if few tracks exist and are therefore unable to perform poorly. PPO-BiGRU+MHSA again performs the best with a mean GOSPA distance of $\SI{1.13E+07}{} \pm \SI{2.32E+06}{}$.
This is due to the lowest GOSPA missed score (\autoref{fig:results_gospa_missed}) of $\SI{1.04E+11}{} \pm \SI{2.98E+10}{}$
even though the GOSPA localisation score (\autoref{fig:results_gospa_localisation}) was the highest of all algorithms with a score of $SI{1.37E+10}{} \pm \SI{9.79E+09}{}$.

\section{Conclusions and Future Work}
\label{sect:conclusion_future_work}



PPO-BiGRU+MHSA exhibited the best overall performance, while the static algorithm, lacking tracking and search capabilities, performed the worst. PPO-BiGRU+MHSA excelled in tracking but not search. Notably, PPO-BiGRU performed similarly to PPO-Flat, highlighting the critical role of the MHSA, which effectively learns the importance of tracks within the list. PPO-Flat, by flattening the track list, burdens the policy without the benefits of an optimized architecture. This study limited actions to radar azimuth and elevation commands, assuming instantaneous action. In practice, platform movement within the area of regard should be considered. The current MTT could be improved to provide time-series trajectory histories as in the trajectory-PMBM \cite{garcia2020trajectory} providing additional information to the agent.

\bibliography{references}

\end{document}